\pdfoutput=1
\documentclass[11pt]{article}
\usepackage{times}
\usepackage{latexsym}

\usepackage{graphicx}
\usepackage{amsmath}
\usepackage{amsfonts}
\usepackage{multirow}
\usepackage{float}
\usepackage{booktabs}
\usepackage{makecell}
\usepackage{multirow}
\usepackage{algorithm}
\usepackage{algpseudocode}
\usepackage{multirow, makecell}
\usepackage{pifont}
\usepackage[]{acl}

\usepackage{times}
\usepackage{latexsym}

\usepackage[T1]{fontenc}
\usepackage[utf8]{inputenc}

\usepackage{microtype}

\title{The YiTrans End-to-End Speech Translation System \\ for IWSLT 2022 Offline Shared Task}

\author{Ziqiang Zhang$^{1}$\thanks{\ \ Equal contributions during internship at Microsoft Research Asia.}, Junyi Ao$^{2 \ast}$, Long Zhou$^{3}$, Shujie Liu$^{3}$, Furu Wei$^{3}$, Jinyu Li$^{3}$ \\
$^1$School of Information Science and Technology, \\ University of Science and Technology of China \\
$^2$School of Data Science, The Chinese University of Hong Kong (Shenzhen) \\
$^3$Microsoft \\
}

\date{}

\begin{document}
\maketitle
\begin{abstract}

This paper describes the submission of our end-to-end YiTrans speech translation system for the IWSLT 2022 offline task, which translates from English audio to German, Chinese, and Japanese.
%text directly in both cascaded and end-to-end conditions.
The YiTrans system is built on large-scale pre-trained encoder-decoder models.
More specifically, we first design a multi-stage pre-training strategy to build a multi-modality model with a large amount of labeled and unlabeled data.
We then fine-tune the corresponding components of the model for the downstream speech translation tasks.
Moreover, we make various efforts to improve performance, such as data filtering, data augmentation, speech segmentation, model ensemble, and so on.
Experimental results show that our YiTrans system obtains a significant improvement than the strong baseline on three translation directions, and it achieves +5.2 BLEU improvements over last year's optimal end-to-end system on tst2021 English-German.
Our final submissions rank first on English-German and English-Chinese end-to-end systems in terms of the automatic evaluation metric.
We make our code and models publicly available.\footnote{\url{https://github.com/microsoft/SpeechT5}}

%and the end-to-end system achieves a comparable performance with the cascaded system on English-German translation.

\end{abstract}

\section{Introduction}

In this paper, we describe our end-to-end speech translation system YiTrans which participates in the offline tracks of the IWSLT 2022 evaluation campaign.
We evaluate our systems from English to German, Chinese and Japanese.
We aim at exploring the pre-training methods for end-to-end systems, and bridging the quality gap with the cascaded approaches.

As self-supervised learning has been shown effective in speech-to-text tasks \cite{baevski2020wav2vec, hsu2021hubert, ao2021speecht5, bapna2021slam}, our teams are interested in building a multi-modality pre-trained model with self-supervised approaches by leveraging large amounts of speech and text data.
% We explore various model architectures for the related tasks.
% to investigate the knowledge transferring between the text to text and speech to text task.
Inspired by SpeechT5 \cite{ao2021speecht5}, we design a multi-stage unified-modal training strategy for pre-training both the encoder and decoder.
Our final end-to-end ST systems are built by fine-tuning the pre-trained models.
%which can boost the performance after fine-tuning for all the tasks, including ASR, MT, and ST.

This paper also tries to improve the system performance by exploring various techniques for the related tasks.
(1) To boost the performance with advanced speech segmentation \cite{anastasopoulos-etal-2021-findings}, we apply the pyannote toolkit \cite{bredin2020pyannote} and the merge algorithm from \citet{inaguma-etal-2021-espnet} to segment the audio. Particularly, to overcome the long sentence problem in the dataset, we design a new segment algorithm.
(2) Dataset is the key point for a ST system to perform well. Hence, we conduct refined data filtering and large-scale data augmentation \cite{jia2019leveraging}.
(3) We also employ progressive learning, back translation and multi-stage fine-tuning \cite{yang2021multilingual,sennrich2015improving,wang-etal-2020-casias} when fine-tuning our models.
% including data filtering, multi-task training, and model ensembling
(4) Motivated by \citet{tang-etal-2021-fst}, we utilize joint ST and MT fine-tuning for our end-to-end ST models.
(5) As comparison, we also build the cascaded systems for all three language pairs by fine-tuning ASR and MT models from pre-trained models.

The rest of this paper is organized as follows.
In Section 2, we describe the data preparation, including the data pre-processing, data augmentation, and speech segmentation.
Section 3 illustrates the unified-modal pre-training methods, and our systems for all three tasks.
We share the experimental setting, results, and analyses in Section 4.
Section 5 concludes the submission.
We also present the official test results \cite{iwslt2022} of our submitted system in Appendix \ref{appendix}.

\section{Data Preparation}

\subsection{Datasets}

Our system is built under constraint conditions.
The training data can be divided into five categories: unlabeled audio, monolingual text, ASR, MT, and ST corpora1.

\begin{table}[!h]
\begin{center}
\begin{tabular}{lcc}
\toprule
Datasets & \# Utterances & \# Hours \\
\midrule
\midrule
\textbf{\textit{Unlabeled Data}} \\
\hline
VoxPopuli & 1224.9k & 28708 \\
\midrule
\midrule
\textbf{\textit{Labeled ASR Data}} \\
\hline
MuST-C v1\&v2 & 341.6k & 616.9 \\
ST-TED & 171.1k & 272.8 \\
LibriSpeech & 281.2k & 961.1 \\
CoVoST & 288.4k & 426.1 \\
CommonVoice & 1224.9k & 1668.1 \\
TEDLIUM v2\&v3 & 361.2k  & 660.6 \\
Europarl & 34.3k & 81.4 \\
VoxPopuli ASR & 177.0k & 501.3 \\
% \hline
% Sum of Labeled Data & 2879.7k & 5188.3 \\
% \midrule
% \midrule
% Sum & 4104.6k & 33896.3 \\
\midrule
\midrule
\textbf{\textit{Labeled ST Data}} \\
\hline
\textbf{en-de} \\
\hline
MuST-C v2 & 249.8k & 435.9 \\
ST-TED & 171.1k & 272.8 \\
CoVoST & 288.4k & 426.1 \\
Europarl & 32.6k & 77.2 \\
\hline
\textbf{en-ja} \\
\hline
MuST-C v2 & 328.4k & 534.5 \\
CoVoST & 288.4k & 426.1  \\
\hline
\textbf{en-zh} \\
\hline
MuST-C v2 & 358.5k & 586.8 \\
CoVoST & 288.4k & 426.1 \\
\bottomrule
\end{tabular}
\end{center}
\caption{\label{audio_stat} English audio data statistics}
\end{table}

\paragraph{Unlabeled Audio} 
We utilize large-scale unlabeled and labeled audio for pre-training. 
As shown in Table \ref{audio_stat}, we pre-train our models by using around 28k hours of unlabeled audio data from VoxPopuli \cite{wang-etal-2021-voxpopuli}, and around 5.1k hours of labeled ASR data, which will be introduced later.

\paragraph{Monolingual Text} 
Monolingual text is used either for pre-training or back-translation.
We collect data for English as well as three target languages from WMT21 news translation task\footnote{https://www.statmt.org/wmt21/translation-task.html}, including News Commentary\footnote{http://data.statmt.org/news-commentary}, Europarl v10\footnote{http://www.statmt.org/europarl/v10}, News crawl\footnote{http://data.statmt.org/news-crawl}, and Common Crawl\footnote{http://data.statmt.org/ngrams}.
As Common Crawl contains much noisier data, it is only used for \textbf{ja} and \textbf{zh} to expand the collected data size to 500M.
The statistics are listed in Table \ref{mono_stat}.

\begin{table}[]
\begin{center}
\resizebox{\linewidth}{!}{
\begin{tabular}{llccc}
\toprule
 & en & de & ja & zh \\
\midrule
\midrule
Collected & 341M & \multicolumn{1}{l}{389M} & \multicolumn{1}{l}{500M} & \multicolumn{1}{l}{500M} \\
\midrule
Processed \& filtered & 50M & 50M & 50M & 50M \\
\bottomrule
\end{tabular}
}
\end{center}
\caption{\label{mono_stat} Monolingual text data statistics}
\end{table}

\paragraph{ASR Corpus} 

For training and evaluation of our ASR models, we use MuST-C v1 \cite{di-gangi-etal-2019-must}, MuST-C v2 \cite{CATTONI2021101155}, ST-TED \cite{niehues-etal-2018-iwslt}, LibriSpeech \cite{Panayotov2015ls}, CoVoST 2 \cite{wang2020covost}, TED-LIUM v2 \cite{rousseau-etal-2012-ted}, TED-LIUM v3 \cite{hernandez2018ted}, Europarl \cite{koehn2005europarl}, VoxPopuli ASR data, and Mozilla Common Voice \cite{ardila2019common}, which results in around 5188.3hr labled ASR data as shown in Table \ref{audio_stat}.
For MuSTC-C and Europarl, we collected the data from all language pairs and removed the overlap audios according to the audio id.

\begin{table}[!h]
\begin{center}
\resizebox{\linewidth}{!}{
\begin{tabular}{lccc}
\toprule
Datasets & en-de & en-ja & en-zh \\
\midrule
\midrule
\textbf{\textit{In-domain}} \\
\hline
MuST-C v2 & 249.8k & 328.4k & 358.5k \\
TED & 209.5k & 223.1k & 231.3k \\
\midrule
\midrule
\textbf{\textit{Out-of-domain}} \\
\hline
CoVoST & 288.4k & 288.4k & 288.4k \\
Europarl & 32.6k & - & - \\
OpenSubtitles2018 & 18.7M & 1.9M & 10.0M \\
WMT21 & \multicolumn{1}{l}{93.3M} & \multicolumn{1}{l}{16.6M} & \multicolumn{1}{l}{61.0M} \\
\hline
Sum (processed) & \multicolumn{1}{l}{82.0M} & \multicolumn{1}{l}{13.8M} & \multicolumn{1}{l}{51.5M} \\
Sum (filtered) & \multicolumn{1}{l}{16.1M} & \multicolumn{1}{l}{3.6M} & \multicolumn{1}{l}{7.6M} \\
\bottomrule
\end{tabular}
}
\end{center}
\caption{\label{mt_stat} MT data statistics}
\end{table}

\paragraph{MT Corpus} 
Machine translation (MT) corpora are used to translate the English transcription.
For training and evaluation of our MT models, we use MuST-C v2 and TED corpus \cite{cettolo2012wit3} as in-domain data.
We also use CoVoST 2, Europarl, OpenSubtitles2018 \cite{lison2016opensubtitles2016} as well as all available paired data provided by WMT21 as out-of-domain data.
The statistics are listed in Table \ref{mt_stat}.

\paragraph{ST Corpus} 
The ST corpus we used includes the MuST-C v2, ST-TED, CoVoST 2 and Europarl, as listed in Table \ref{audio_stat}.
MuST-C v2 and ST-TED are treated as in-domain data. The ST corpus can be greatly expanded by large-scale data augmentation, which will be introduced in the following Section.

\subsection{Text Processing \& Filtering}
For monolingual and out-of-domain MT data, we first process the text through the following steps:

(1) We clean up the data by removing sentences that have non-printing characters, http tags or words with length longer than 50 characters (words are separated by space, for \textbf{ja} and \textbf{zh} the threshold is 150).
The processed text data is then deduplicated.

(2) We use fast-text \footnote{https://github.com/facebookresearch/fastText} \cite{joulin2016fasttext} to filter out the sentences with invalid languages.

(3) For paired data, we use fast\_align\footnote{https://github.com/clab/fast\_align} \cite{dyer2013simple} to calculate the alignment quality, which is evaluated by the percentage of aligned words.
We remove 20\% of data with the lowest alignment quality.

(4) We then use XenC\footnote{https://github.com/antho-rousseau/XenC} \cite{rousseau2013xenc} to perform domain filtering.
It computes the distinction of two n-gram language models, which are in-domain and out-of-domain language models.
The amount of selected data is 50M for monolingual text, and for paired text it depends on the XenC scores.
The results are listed in Table \ref{mono_stat} and \ref{mt_stat}.

\subsection{Post processing}
We only do post-processing for \textbf{en-ja} systems as an optional choice.
It is because we noticed that for \textbf{en-ja} there is few punctuations in the target side of training data.
To obtain translation results with rich punctuation, which are more natural in the real world, we train a punctuation model to post-process the translated results.
The model is initialized from mBART50 \cite{tang2020multilingual} and trained to predict sentences with proper punctuation.
The training data is collected from out-of-domain \textbf{en-ja} MT data.
We select the sentences with rich punctuation in Japanese side.

\subsection{Data Augmentation}
The quality of end-to-end ST is often limited by a paucity of training data, since it is difficult to collect large parallel corpora of speech and translated transcript pairs
In this paper, we attempt to build a large amount of synthetic data for ST and MT, separately. We will introduce the data augmentation method in Section \ref{system_description} in detail.

\subsection{Speech Segmentation}

\begin{algorithm}
\footnotesize
\caption{Segment audios based on pyannote toolkit}\label{alg:segment}
\begin{algorithmic}[1]
\Function{SegmentAudio}{$x$, $P_{on}$, $P_{off}$, $T_{dur}$}
\State $L \gets VAD(x, P_{on}, P_{off})$ \Comment{$\{a_1,$ $ ...,$ $a_n\}$}
\State $L_{new} \gets \{ \}$
\For{$a_i  \in  L$}
    \If{$a_i.length > T_{dur}$}
        \If{$P_{on} < 0.95$ or $P_{off} < 0.95$}
            \State $L_{new} \gets L_{new} \cup $ \Call{SegmentAudio}{$a_i$, $P_{on}+\alpha_{on}$, $P_{off} +\alpha_{off}$, $T_{dur}$}
        \Else{}
            \State $L_{new} \gets L_{new} \cup $ \Call{EqualSegment}{$a_i$}
        \EndIf
    \EndIf
\EndFor
\State \Return {$L_{new}$}
\EndFunction
\end{algorithmic}
\end{algorithm}

Similar to the previous evaluation, this year’s evaluation data are segmented using an automatic tool, which does not ensure that segments are proper sentences nor that they are aligned with the translated text.
In addition, there is an apparent mismatch for segmentation between using voice activity detection (VAD) and segmenting by punctuations, where the latter is usually used for segmenting the training data.
These assign extra importance to develop methods for proper segmentation of the audio data, which was confirmed in the previous year’s evaluation campaign, where all top submissions used their own segmentation algorithm \cite{anastasopoulos-etal-2021-findings}.
% Therefore, it is crucial to design a better segmentation algorithm.

Therefore, we design a segmentation algorithm based on a VAD model provided by pyannote.audio\footnote{https://huggingface.co/pyannote/voice-activity-detection} \cite{bredin2020pyannote}, as illustrated in Algorithm \ref{alg:segment}.
We find that long segments are difficult for the model to decode and need to be further segmented.
More specifically, we firstly use the VAD model pre-trained on AMI dataset \cite{Carletta2007UnleashingTK} to segment the audio. 
Two hyperparameters, $P_{on}$ and $P_{off}$, are set for the VAD model, which are the onset speaker activation threshold and offset speaker activation threshold, respectively.
Then the segments longer than $T_{dur}$ are further segmented by increasing $P_{on}$ and $P_{off}$ with $\alpha_{on}$ and $\alpha_{off}$ if $P_{on}$ and $P_{off}$ are smaller than $0.95$.
Otherwise, we segment the audio into several parts with the same length smaller than $T_{dur}$, as large activation thresholds may lead to incorrect segmentation.
In our experiments, We use the default values of the pre-trained model for $P_{on}$ and $P_{off}$, which are $0.481$ and $0.810$. respectively.
For segmenting long audios, we set the $T_{dur}$ to 43.75 seconds, $\alpha_{on}$ to $0.1$, and $\alpha_{off}$ to $0.028$.

Moreover, some short segments are generated by the VAD model according to our observations, which may be incomplete sentences and harm the performance of our ST model.
Merging the short segments helps the ST model utilize the context information.
So we follow the algorithm in \cite{inaguma-etal-2021-espnet} to merge the short segments after the segmentation.

\section{End-to-End YiTrans ST System}
\label{system_description}

%\subsection{Large-Scale Pre-Training for E2E ST}

Recent studies, such as SpeechT5 \cite{ao2021speecht5} and SLAM \cite{bapna2021slam}, have shown that joint pre-training of speech and text can boost the performance of spoken language processing tasks, such as speech translation.
% Recent studies show that the fully end-to-end solution achieves promising performance. 
This section will mainly introduce the model architecture of our end-to-end YiTrans system, and the proposed methods to pre-train and fine-tune the models.

\subsection{Model Architecture}
\begin{figure}[t]
	\centering
	\setlength{\abovecaptionskip}{3pt}
    \setlength{\belowcaptionskip}{-0pt}
	\includegraphics[width=\linewidth]{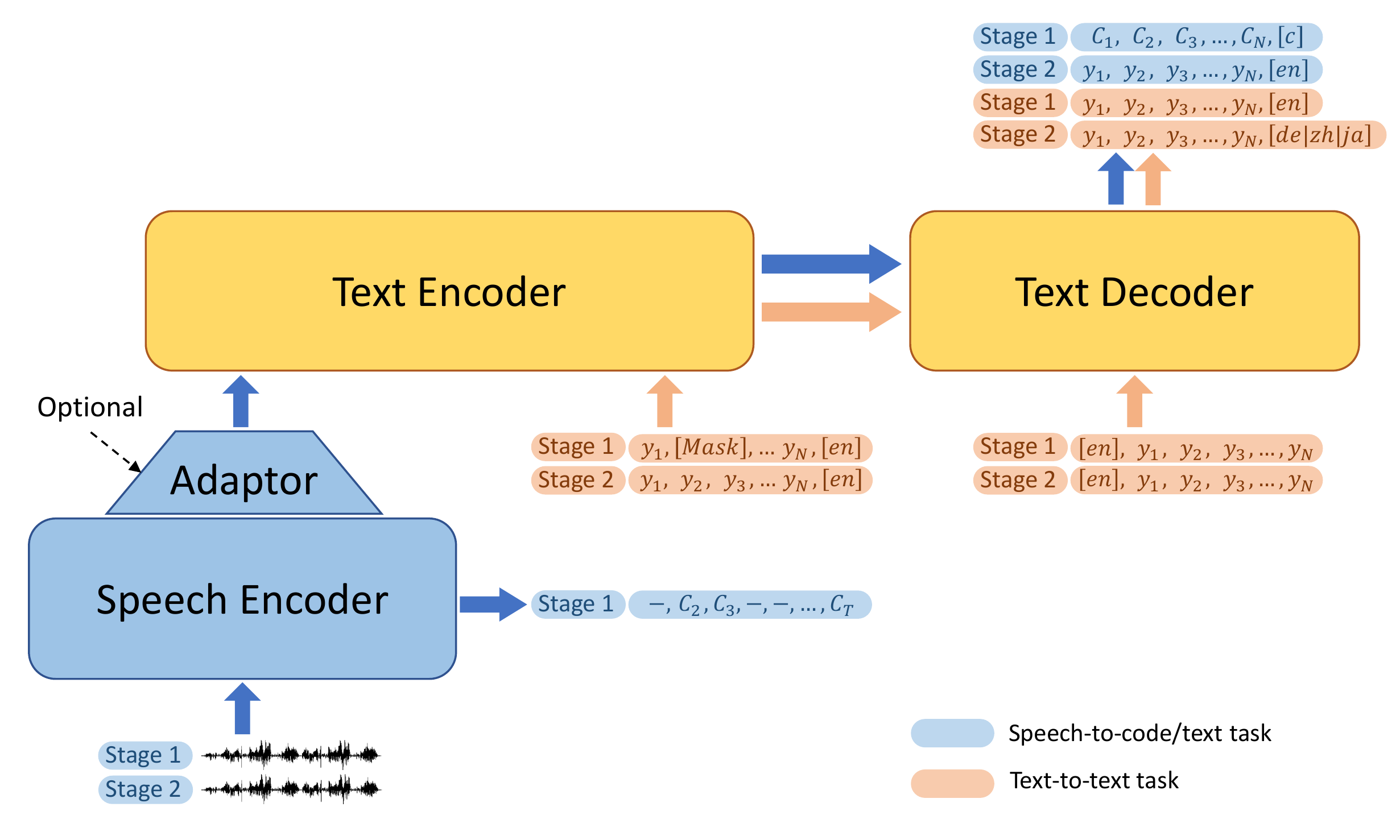}
	\caption{An illustration of the pre-training model.}\label{architecture}
\vspace{-12pt}\end{figure}

%Our system is based on deep Transformerimplemented on the fairseq toolkit.
Our evaluation system is based on an encoder-decoder model with state-of-the-art Transformer architecture. Figure \ref{architecture} shows the framework of our end-to-end speech translation model, which consists of a speech encoder, text encoder, and text decoder. We employ the relative positional encoding \cite{shaw-etal-2018-self} for both the encoder and decoder network.

The speech encoder network contains a convolutional feature encoder and a Transformer encoder.
The convolutional feature encoder is a convolutional network for extracting feature from waveform, which has seven 512-channel layers with kernel widths [10,3,3,3,3,2,2] and strides [5,2,2,2,2,2,2].
The Transformer encoder has 24 layers with model dimension 1024, inner dimension 4096 and 16 attention heads.
The text encoder and decoder contain 12 layers and have a similar architecture to the Transformer encoder, except that the text decoder includes the cross-attention and the masked self attention.
We optionally add an adaptor between the speech encoder and text encoder, which is three one-dimensional convolution layers with stride 2.

\subsection{Multi-Stage Unified-Modal Pre-Training}
% Previous work, such as SpeechT5 \cite{ao2021speecht5} and SLAM \cite{bapna2021slam}, have shown that joint pre-training of speech and text can boost the performance of spoken language processing tasks, such as ASR and ST.
To leverage large amounts of speech and text data, we firstly initialize the speech encoder with the HuBERT \textsc{Large} \cite{hsu2021hubert} and the text encoder and decoder with the mBART50 \cite{tang2020multilingual}.
Then we design a multi-stage pre-training strategy to boost the performance of ASR and ST tasks.

% To make full use of unlabeled speech data, following Speech2C \cite{ao2022speech2c}, 

In the first stage, we employ the speech to code pre-training method following Speech2C \cite{ao2022speech2c} to make full use of unlabeled speech data.
More specifically, We set two pre-training tasks for the encoder-decoder pre-training using unlabeled speech data with pseudo codes, which are acoustic units learned from an offline clustering model.
The encoder of Speech2C predicts the pseudo code via masked language modeling (MLM) in encoder output, like HuBERT model.
In addition to MLM loss, the decoder of Speech2C learns to reconstruct pseudo codes auto-regressively, instead of generating real text transcription, both of which are discrete representations and have some semantic information corresponding to the speech signal.
For the text data, the BART loss \cite{lewis-etal-2020-bart} and cross entropy loss are used for the monolingual English data and MT data of three target languages, respectively.
Note that the text data is only used for pre-training the text encoder and text decoder. For the second stage, we use the ASR data and the filtered MT data to continuously pre-train the model. 

% \paragraph{Joint Pre-Training of Speech and Text}
% Previous work, such as SpeechT5 \cite{ao2021speecht5} and SLAM \cite{bapna2021slam}, have shown that joint pre-training of speech and text can boost the performance of spoken language processing tasks, such as automatic speech recognition and speech translation.

% (1) unlabeled speech and text

% (2) ASR data and MT data

\subsection{Joint Fine-Tuning}
After pre-training, all the pre-trained modules (speech encoder, text encoder, text decoder and the optional adaptor) are used for directly fine-tunig a end-to-end ST model.
We also make various efforts to improve the final perfermance.

\paragraph{Joint ST and MT Fine-Tuning}
We train the ST model along with an auxiliary text to text machine translation (MT) task. 
We utilize two methods from \cite{tang2021improving} to enhance the performance of the primary ST task.
First, a cross-attentive regularization is introduced for the encoders. It minimizes the L2 distance between two reconstructed encoder output sequences and encourages the encoder outputs from different modalities to be closer to each other. 
Second, online knowledge distillation learning is introduced for MTL in order to enhance knowledge transfer from the MT to the ST task.

\paragraph{Synthetic Data for ST}

To provide more parallel audio-translation pairs,
we translate the English side of the ASR data with our MT model.
Specifically, we translate all the transcriptions of labeled ASR data listed in Table \ref{audio_stat} to three target languages.
For \textbf{en-de}, we additionally generate a certain amount of (about 8000 hours) cascaded pseudo data from unlabeled VoxPopuli, by firstly generating pseudo transcriptions with ASR model and then translating them with MT model.

\paragraph{Multi-Stage Fine-Tuning}
Note that our ST data is from various domains, including synthetic data and out-of-domain data (e.g. CoVoST).
To make out ST model better adapted to the TED domain, we adopt the multi-stage fine-tuning method 
according to data category:
At the first stage, we fine-tune ST models with all ST data, including synthetic and true data;
Then at the second stage, the ST models are continually fine-tuned with in-domain data, i.e. Must-C and ST-TED.

% \subsubsection{Data Augmentation}

% Data augmentation has been shown to provide
% increased performance in both ASR and ST, by enriching and diversifying the training data.

\subsection{Cascaded Speech Translation}

To compare with our end-to-end YiTrans system, we also build a cascaded system by fine-tuning ASR and MT models from pre-trained models, and these subsystems also has been used to construct synthetic data for ST.

\subsubsection{Automatic Speech Recognition}
%

% For ASR, we fine-tune the model from the second stage of pre-training with all of the ASR data.
% We train a punctuation model using the English text of the MuST-C dataset, and add punctuations to the transcriptions of the TEDLIUM and LibriSpeech dataset with this model.
% To let the model fit the TED domain, we also fine-tune the model with only the MuST-C dataset and TED-style datasets.
% The TED-style datasets include MuST-C, ST-TED, and TED-LIUM corpus.

We fine-tune our ASR model with the following strategies:
(1) \textbf{Synthetic Data for ASR}.
To make the transcriptions contain the punctuations, we train a punctuation model using the English text of the MuST-C dataset, and add punctuations to the transcriptions of the TEDLIUM and LibriSpeech dataset with this model.
We also use a model trained on MuST-C dataset to synthesize data from the Voxpopuli corpus.
(2) \textbf{Data Filtering}.
We find that the ASR data contains some noise and the transcription of some utterances are wrong.
Therefore, we also use a model trained on MuST-C dataset to calculate the WER of each sentence, which is used for filtering ASR data.
(3) \textbf{In-Domain Fine-Tuning}.
To let the model fit the TED domain, we train two models from the second stage of pre-training.
For the first one, we directly fine-tune the model on the MuST-C dataset.
For the second one, we train the model with the TED-style datasets, which include MuST-C, ST-TED, and TED-LIUM corpus.
We also filter the utterances that the WER is larger than 50\% for the second model.

\subsubsection{Machine Translation}

All of our MT models for the offline task are fine-tuned from the big pre-trained mBART50 model, with advanced techniques:
% (1) \textbf{Progressive Learning}. \label{ssec:progressive_learning}
%To take advantage of pre-trained model as well as a large amount of MT data, we inherit the idea of progressive training \cite{li-etal-2020-shallow}, which trains the model from shallow to deep.
(1) We inherit the idea of \textbf{Progressive Learning} \cite{li-etal-2020-shallow} \label{ssec:progressive_learning} to train the model from shallow to deep.
Specifically, our MT model has 24 encoders and 12 decoder layers, where the top 12 encoder layers are randomly initialized and the rest layers are initialized from mBART50.
(2) \textbf{Back Translation}.
%Back-translation is an effective way to improve the translation quality by leveraging a large amount of monolingual data and has been widely used in WMT evaluation campaigns \cite{akhbardeh-etal-2021-findings}.
Following previous experience in WMT evaluation campaigns \cite{akhbardeh-etal-2021-findings}, we use the trained \{\textbf{de},\textbf{ja},\textbf{zh}\}-\textbf{en} MT models to generate the English side for the selected monolingual text from Table \ref{mono_stat}. The MT models are also fine-tuned form mBART50.
All back-translated pairs and the true paired data are combined for training.
(3) \textbf{Multi-Stage Fine-Tuning}.
We also perform multi-stage fine-tuning for MT models, where the model is first fine-tuned with all (processed) MT data, then is fine-tuned with in-domain data for a few steps.
There is also an optional stage between them, which is fine-tuning with in-domain filtered data (the last line in Table \ref{mt_stat}).
(4) \textbf{ASR Output Adaptation}.
To alleviate the mismatch between the ASR transcripts and the real text used for training MT models, we 
add the synthetic in-domain data at the in-domain fine-tuning stage.
The synthetic data is generated by replacing the English site text with pseudo ASR labels.

% \paragraph{Ensemble Decoding}

% \paragraph{Merge Algorithm}
% We introduce a merge algorithm to combine short sentences to long sentences.

%different segment method

%our merge method

\section{Experiments \& Results}

\subsection{Pre-Training Setup}

All models are implemented in Fairseq \footnote{https://github.com/pytorch/fairseq} \cite{ott2019fairseq}.
We pre-train two models depending on the computational efficiency. 
The first has 24 speech encoder layers, 12 text encoder layers and 12 decoder layers (denoted as PT48).
The second has 12 encoder layers, an adaptor, 12 text encoder layers and 12 decoder layers (denoted as PT36).
The total number of parameters for the pre-trained model is about 927M and 803M,  respectively.
The vocabulary size is 250k, which is inherited from the mBART50 model.

For the first stage, we pre-train our model on 64 A100 GPUs with a batch size of 37.5s samples per GPU for speech and 1875 tokens per GPU for text and set the update frequency to 3 for 100k steps.
We optimize the model with Adam \cite{kingma2014adam} and set the learning rate to 3e-5, which is warmed up for the first 8\% of updates and linearly decayed for the following updates.
For the second stage, we also use 64 A100 GPUs and train the model for 300k with a batch size of 30s samples per GPU for speech and 1500 tokens for text.
The learning rate set to 3e-5 is warmed up for the first 10\% steps, held as a constant for the following 40\% steps, and is decayed linearly for the rest steps.
We add a language ID symbol for four languages at the start of each sentence.

\begin{table}[!htp]
\centering
\resizebox{\linewidth}{!}{
\begin{tabular}{clcc}
\hline
ID & Model                                      & tst2019 & tst2020 \\ \hline
1  & Hubert \& mBART                            & 30.72   & 31.58   \\
2  & \quad + in-domain FT                       & 30.62   & 33.07   \\ \hline
3  & PT36 + joint FT                            & 20.10 (*)   & 20.12 (*)   \\
4  & \quad + in-domain FT                       & 30.01   & 32.65   \\ \hline
5  & PT48                                       & 30.56   & 33.26   \\
6  & \quad + in-domain FT                       & 30.98   & 33.48   \\
7  & \quad + joint FT                           & 30.65   & 33.16   \\
8  & \quad \quad + in-domain FT                 & \textbf{31.02}   & 33.46   \\
9  & \quad + cascaded data                      & 31.00   & \textbf{33.52}   \\
10 & \quad \quad + in-domain FT                 & 30.91   & 33.42   \\ \hline
11 & Ensemble (10, 6)                           & 31.46   & 34.03   \\
12 & Ensemble (10, 8, 6)                        & 31.49   & 33.84   \\
13 & Ensemble (10, 9, 8, 6)                     & 31.47   & 33.95   \\
14 & Ensemble (10, 9, 8, 6, 2)                  & \textbf{31.57}   & 33.96   \\
15 & Ensemble (10, 9, 8, 6, 4, 2)               & 31.40   & \textbf{34.10}   \\ \hline
\end{tabular}
}
\caption{\label{exp_e2e_de} BLEU results of e2e \textbf{en-de} models.}
\end{table}

\begin{table}[!htp]
\centering
\small
% \resizebox{\linewidth}{!}{
\begin{tabular}{clc}
\hline
 & Model                       & tst-common \\ \hline
1 & Hubert \& mBART             & 18.13      \\
2 & \quad + in-domain FT        & 18.59      \\ \hline
3 & PT36 + joint FT             & 18.16      \\
4 & \quad + in-domain FT        & 18.86      \\ \hline
5 & PT48                        & 17.67      \\
6 & \quad + in-domain FT        & 18.30      \\
7 & \quad + joint FT            & 18.71      \\
8 & \quad \quad + in-domain FT  & \textbf{19.13}      \\ \hline
9 & Ensemble (8, 6)             & 19.38      \\
10 & Ensemble (8, 6, 2)         & 19.48      \\
11 & Ensemble (8, 6, 4)         & 19.70      \\
12 & Ensemble (8, 6, 4, 2)      & \textbf{19.81}      \\ \hline
\end{tabular}
% }
\caption{\label{exp_e2e_ja} BLEU results of e2e \textbf{en-ja} models.}
\end{table}

\subsection{End-to-End Speech Translation}
Our e2e ST models are fine-tuned from various pre-trained models.
When fine-tuning with all ST data, the learning rate is set to 5e-5 and then is decayed linearly to zero within 200k training steps.
And when fine-tuning with in-domain data, the learning rate is set to 1e-5 for 30k steps.
All ST models are fine-tuned on 8 A100 GPUs with a batch size of about 30s per GPU and update frequency of 4.
\begin{table}[!htp]
\centering
\small
% \resizebox{\linewidth}{!}{
\begin{tabular}{clc}
\hline
 & Model                       & tst-common \\ \hline
1 & Hubert \& mBART             & 28.69      \\
2 & \quad + in-domain FT        & 28.71      \\ \hline
3 & PT36                        & 28.62      \\
4 & \quad + in-domain FT        & 28.61      \\ \hline
5 & PT48                        & 29.07      \\
6 & \quad + in-domain FT        & \textbf{29.26}      \\
7 & \quad + joint FT            & 28.51      \\
8 & \quad \quad + in-domain FT  & 29.14      \\ \hline
9 & Ensemble (8, 6)             & 29.38           \\
10 & Ensemble (8, 6, 4)         & 29.36           \\
11 & Ensemble (8, 6, 2)         & 29.48           \\
12 & Ensemble (8, 6, 4, 2)      & \textbf{29.53}           \\ \hline
\end{tabular}
% }
\caption{\label{exp_e2e_zh} BLEU results of e2e \textbf{en-zh} models.}
\end{table}

\begin{table*}[!htp]
\small
\begin{center}
\begin{tabular}{@{\extracolsep{2pt}}lcccc@{}}
    \toprule
    \multirow{2}{*}{Model} & en-de & en-ja/zh   & \multirow{2}{*}{tst2019} & \multirow{2}{*}{tst2020} \\
    &  tst-common & tst-common & & \\
    \midrule
    \midrule
      Fine-tune with TED-Style data  & 8.49 & 8.67 & 10.9 & 13.4  \\
      Fine-tune with MuST-C  & 8.55 & 8.70 & 10.9 & 13.6  \\
    \midrule
      ensemble & 8.47 & 8.56 & 10.7 &  13.3 \\
    \bottomrule
\end{tabular}
\end{center}
\caption{\label{wer_asr} WER results of ASR Systems.}
\end{table*}

\begin{table*}[!htp]
\centering
\small
% \resizebox{\linewidth}{!}{%
\begin{tabular}{clc|c|c|c}
\hline
 & \multirow{2}{*}{Method} & \multirow{2}{*}{Model size} & MT en-de & MT en-ja & MT en-zh \\
 & &  & tst-common & tst-common & tst-common \\ \hline
1 & Baseline                    & 12-12     & 35.82     & 19.58     & 28.52 \\
2 & \quad + in-domain FT        & 12-12     & 37.01     & 20.21     & 30.10 \\ \hline
3 & Deep model                  & 24-12     & 36.25     & 20.15     & 29.19 \\
4 & \quad + data filtering      & 24-12     & 37.38     & 24.52 (*) & 29.22 \\
5 & \quad \quad + in-domain FT  & 24-12     & \textbf{38.27}     & \textbf{24.91} (*) & 29.94 \\ \hline
6 & Back-translation            & 24-12     & 37.29     & 18.62     & 28.65 \\
7 & \quad + in-domain FT        & 24-12     & 38.05     & 20.92     & \textbf{30.43} \\ \hline
\end{tabular}%
% }
\caption{\label{exp_mt} BLEU results of MT systems. * indicates the results may be over-fitted on tst-common set.}
\end{table*}

\paragraph{en-de}
We use \textit{tst2019} and \textit{tst2020} as validation sets.
We do not use \textit{tst-common} as we find that it has overlapped speech samples with ST-TED training data.
All BLEU results are computed at paragraph level, as listed in Table \ref{exp_e2e_de}.
It is noticed that almost all of the models get improved when fine-tuned with in-domain data (in-domain FT).
What's more, joint ST\&MT fine-tuning (joint FT) and adding cascaded pseudo ST data also help the performance.
%Table \ref{exp_e2e_de} shows that
While PT36 fine-tuned models get some unexpectedly bad results without in-domain fine-tuning.
After checking the results we found that sometimes the model could only be able to decode a small portion of a sample especially when the sample is long.
Finally, our PT48 fine-tuned model achieves the best performance, and ensemble decoding \cite{liu2018comparable} with different models continually brings improvement.
Our final submitted system is the last line of Table \ref{exp_e2e_de}.

\paragraph{en-ja}
We use \textit{tst-common} as the validation set.
%, which has sentence-level translations so that BLEUs are computed at the sentence level.
The results are listed in Table \ref{exp_e2e_ja}, where the BLEUs are computed after tokenized by Mecab\footnote{https://taku910.github.io/mecab/}.
Cascaded pseudo ST data is not performed due to the time urgency.
Similar phenomena could be observed in Table \ref{exp_e2e_ja}, where in-domain fine-tuning, joint ST\&MT fine-tuning as well as model ensemble benefit the translation performance.
Again, our PT48 fine-tuned model achieves the best performance.
Our submitted system are listed in the last line of Table \ref{exp_e2e_ja}.

\paragraph{en-zh}
The validation set is also \textit{tst-common} and sentence level BLEUs with character  tokenizer are reported in Table \ref{exp_e2e_zh}.
We find that in-domain fine-tuning and joint ST\&MT fine-tuning are not as effective here as that in \textbf{en-de} and \textbf{en-ja}.
That might be due to the specific data property of \textbf{en-zh}, e.g. all ST data is not mismatched very much with in-domain data.
Finally, PT48 fine-tuned models still achieve the best performance and model ensemble brings improvement.
Our final submitted system are listed in the last line of Table \ref{exp_e2e_zh}.
Note that the results in Table \ref{exp_e2e_zh} are not post-processed, while in our submitted results of  \textit{tst2022}, we post-process the decoding results by correcting the punctuation to Chinese style.

\subsection{Cascade Speech Translation}
\paragraph{Automatic Speech Recognition}
\begin{table*}[!htp]
\centering
\small
% \resizebox{\linewidth}{!}{%
\begin{tabular}{llc|ccc|c|c}
\hline
\multirow{2}{*}{ID} & \multirow{2}{*}{Method} & \multirow{2}{*}{Model size} & \multicolumn{3}{c|}{en-de} & en-ja & en-zh \\
 & & & tst-common & tst2019 & tst2020 & tst-common & tst-common \\ \hline
1 & Baseline & 12-12 & 33.07 & 30.47 & 32.96 & 18.79 & 27.50 \\
2 & \quad + in-domain FT & 12-12 & 34.17 & 31.12 & 33.71 & 19.40 & 28.76 \\ \hline
3 & Deep model & 24-12 & 33.29 & 30.67 & 33.14 & 19.00 & 27.81 \\
4 & \quad + data filtering & 24-12 & 34.65 & 31.34 & 33.85 & 22.77 (*) & 27.99 \\
5 & \quad \quad + in-domain FT & 24-12 & \textbf{35.42} & 31.63 & \textbf{34.29} & \textbf{23.45}  (*) & 28.65 \\ \hline
6 & Back-translation & 24-12 & 34.54 & 31.10 & 33.57 & 17.61 & 27.44 \\
7 & \quad + in-domain FT & 24-12 & 35.40 & \textbf{31.72} & 34.16 & 19.94 & \textbf{29.12} \\ \hline
\end{tabular}%
% }
\caption{\label{exp_cas} BLEU results of cascaded systems. * indicates the results may be over-fitted on tst-common set.}
\end{table*}

\begin{table}[!htp]
\small
\begin{center}
\begin{tabular}{@{\extracolsep{2pt}}ccccc@{}}
    \toprule
     VAD & $M_{dur}$(s) & $M_{int}$(s) & tst2019 & tst2020 \\
    \midrule
    \midrule
    \multirowcell{1}{Given}  & - & - & 26.2 & 27.3 \\
    \midrule
    \multirowcell{9}{pyannote}  & - & - & 15.7 & 16.3  \\
    & 20 & 1 & 11.2 & 14.5 \\
    & 25 & 0.5 & 12.4 & 15.0 \\
    & 25 & 1 & 11.0 & 14.4 \\
    & 25 & 1.5 & 11.6 & 14.3 \\
    & 30 & 0.5 & 12.4 & 14.9 \\
    & 30 & 1 & 10.9 & 14.0 \\
    & 30 & 1.5 & 11.1 & 14.3 \\
    & 35 & 1 & 11.4 & 14.0 \\
    \midrule
    \multirowcell{1}{Algo \ref{alg:segment}}  & 30 & 1 & \textbf{10.9} & \textbf{13.6}  \\
    \bottomrule
\end{tabular}
\end{center}
\caption{\label{exp_seg} Comparison of segmentation ways and merge algorithm for ASR in terms of WER score.}
\end{table}
For the ASR fine-tuning, we use the CTC and cross-entropy loss to train the model \cite{shinji2017hybrid}.
The loss weights are are set to 0.5 for both of them.
We fine-tune the model on 8 A100 GPUs with the update frequency 4 for 120k steps, and set the batch size to around 30s samples per GPU.
The learning rate set to 3e-5 is scheduled with the same strategy as the stage 2 of pre-training.

As shown in Table \ref{exp_seg}, we investigate the impact of speech segmentation with the model fine-tuned on MuST-C dataset.
The pyannote toolkit improve the performance significantly compared to the given segmentation.
The merge algorithm from \citet{inaguma-etal-2021-espnet} further decreases the WER.
We adjust two parameters of merge algorithm, $M_{dur}$ and $M_{int}$.
$M_{dur}$ means the maximum duration after merging, and $M_{int}$ is the minimum interval of two segments that will be merged.
The experiments show that when $M_{dur}$ and $M_{int}$ are set to 30s and 1s, respectively, the model achieves the best performance.
We then apply our Algorithm \ref{alg:segment} to further segment the utterance longer than 43.75s, and the final WERs are 10.9 for tst2019 set and 13.6 for tst2020 set.
Table \ref{wer_asr} shows the WER scores of two ASR systems.
We ensemble these two models and use the results for the cascade system.

\paragraph{Machine Translation}
For all three language pairs, we fine-tune both base models (with 12 encoder layers) and deep models (with 24 encoder layers) as described in Section \ref{ssec:progressive_learning}.
All models are fine-tuned on 8 A100 or V100 GPUs with a batch size of 2048 tokens per GPU, the update frequency is 1.
The learning rate is set to 1e-4 with 5k warming up steps, then it is linearly decayed to zero in total 200k steps.
In case of using additional back-translated data, we set the total training step to 300k.
For in-domain fine-tuning, we only change the learning rate to 1e-5 and the total training step to 30k.

The results of MT systems are shown in Table \ref{exp_mt}.
All BLEUs are computed the same way as e2e ST systems.
Similar to e2e ST results, in-domain fine-tuning (in-domain FT) benefits all MT models.
Progressive learning with deeper models also outperforms their baselines for all languages (line 3 vs. line 1).
While, data filtering is shown effective for \textbf{en-de} but slightly negative for \textbf{en-zh}, which might because we remain too little data for \textbf{en-zh} to train such big models.
It is also noticed that \textbf{en-ja} gets un-normal improvement from filtered data (indicated by *), we speculate data filtering might allow us to collect too similar text to \textit{tst-common} to make the model overfit.
Finally, back translation is shown benefit to all languages (line 7), while for \textbf{en-de} it falls slightly behind the best results, probably because of the amount of paired data already sufficient.

\begin{table}[!htp]
\small
\centering
% \resizebox{\linewidth}{!}{%
\begin{tabular}{lccc}
\hline
Ensembled Models & tst-common & tst2019 & tst2020  \\
\hline
\textit{\textbf{en-de}} \\
\hline
MT \#5; ST \#10 & \textbf{36.44} & \textbf{31.90} & \textbf{34.60} \\
MT \#5,\#7; ST \#10 & 36.31 & 31.89 & 34.60 \\
MT \#5,\#7,\#4; ST \#10 & 36.16 & 31.90  & 34.45 \\
\hline
\textit{\textbf{en-ja}} \\
\hline
*MT \#5; ST \#8 & 22.79 & \textbackslash & \textbackslash \\
*MT \#5,\#4; ST \#8 & \textbf{23.26} & \textbackslash & \textbackslash \\
*MT \#5,\#4,\#7; ST \#8 & 22.97 &  \textbackslash & \textbackslash \\
MT \#7; ST \#8 & 20.02 & \textbackslash & \textbackslash \\
MT \#7,\#2; ST \#8 & 20.12 & \textbackslash & \textbackslash \\
MT \#7,\#2,\#3; ST \#8 & \textbf{20.45} &  \textbackslash & \textbackslash \\
\hline
\textit{\textbf{en-zh}} \\
\hline
MT \#7; ST \#6 & 29.38 & \textbackslash &  \textbackslash\\
MT \#7,\#2; ST \#6 & \textbf{29.48} & \textbackslash & \textbackslash \\
MT \#7,\#2,\#5; ST \#6 & 29.32 & \textbackslash & \textbackslash \\
\hline

\end{tabular}%
% }
\caption{\label{ensemble_cas} BLEU results of cascaded systems. * indicates the results may be over-fitted on tst-common set.}
\end{table}

\paragraph{Cascade Systems}
Cascade systems are built upon ASR and MT systems.
Table \ref{exp_cas} shows the cascade ST results when applying the MT model listed in Table \ref{exp_mt} to our best ASR systems.
It is shown that better MT models always lead to better ST results.
To leverage the end-to-end ST models, we also explore the ensemble of MT and end-to-end ST models as shown in Table \ref{ensemble_cas}.
For \textbf{en-ja}, since the BLEU results of MT model \#4 and \#5 may be over-fitted on tst-common set, we also choose another three models for the ensemble.

\section{Conclusion}
In this paper we describe our End-to-End YiTrans speech translation system for IWSLT 2022 offline task.
We explore building ST systems from large-scale pre-trained models.
Our proposed multi-stage pre-training strategy allows the model to learn multi-modality information from both labeled and unlabeled data, which further improves the performance of downstream end-to-end ST tasks.
Our systems are also built on several popular methods such as data augmentation, joint fine-tuning, model ensemble, and so on.
Massive experiments demonstrate the effectiveness of our introduced methods, and show that the end-to-end YiTrans achieves comparable performance with the strong cascade systems and outperforms the last year's best end-to-end system by 5.2 BLEU in terms of English-German tst2021 set.
%, which shows good prospects for end-to-end speech translation.

\section*{Acknowledgments}

We would like to thank Chengyi Wang, Yu Wu, Shuo Ren, Jian Xue, Peidong Wang, and Yashesh Gaur for valuable discussion and suggestions.

\bibliographystyle{acl_natbib}
\bibliography{acl}

% \clearpage
% \newpage

\appendix
\section{Appendix} \label{appendix}
We present the official test results for our submitted systems.
For \textbf{en-de}, our end-to-end system achieves comparable performance with the cascade system, even the cascaded system is the ensemble of end-to-end and cascaded models.
We also outperforms the best result of the last year by a great margin, especially for end-to-end systems.
For \textbf{en-zh}, the gap between end-to-end and cascaded systems is also small (less than 1 point).
While for \textbf{en-ja} cascaded systems performs better than end-to-end systems, probably because the 
end-to-end and cascaded models are complementary and resulting in a better ensemble.
Meanwhile, it is noticed that adding punctuation for \textbf{en-ja} results is beneficial for \textit{ref2} while harmful for \textit{ref1}.

\begin{table}[bh]
\centering
\resizebox{\linewidth}{!}{%
\begin{tabular}{lccc}
\toprule
Model       & BLEU ref2 & BLEU ref1 & BLEU both \\
\midrule
\midrule
Cascaded    & 25.6      & 23.7      & 36.4     \\ \midrule
E2E YiTrans  & 25.7      & 23.6      & 36.5      \\
\bottomrule
\end{tabular}
}
\caption{\label{official_ende} Official results of our submitted \textbf{en-de} ST systems on tst2022.}
\end{table}

\begin{table}[htb]
\centering
\resizebox{\linewidth}{!}{%
\begin{tabular}{lccc}
\toprule
Model       & BLEU ref2 & BLEU ref1 & BLEU both \\
\midrule
\midrule
\textit{Cascaded} \\ \midrule
IWSLT21 rank-1  & 24.6  & 20.3      & 34.0      \\
The submission    & 28.1      & 23.2      & 39.0     \\ \midrule
\midrule
\textit{End-to-end} \\ \midrule
IWSLT21 rank-1  & 22.6  & 18.3  & 31.0  \\
Our YiTrans  & 27.8      & 23.1      & 38.8      \\
\bottomrule
\end{tabular}
}
\caption{\label{official_ende21} Official results of our submitted \textbf{en-de} ST systems on tst2021.}
\end{table}

\begin{table}[htb]
\centering
\resizebox{\linewidth}{!}{%
\begin{tabular}{lccc}
\toprule
Model       & BLEU ref2 & BLEU ref1 & BLEU both \\
\midrule
\midrule
Cascaded    & 34.7      & 35.0      & 42.9     \\ \midrule
E2E YiTrans  & 34.1      & 34.6      & 42.3      \\
\bottomrule
\end{tabular}
}
\caption{\label{official_enzh} Official results of our submitted \textbf{en-zh} ST systems on tst2022.}
\end{table}

\begin{table}[h!]
\centering
\resizebox{\linewidth}{!}{%
\begin{tabular}{lccc}
\toprule
Model       & BLEU ref2 & BLEU ref1 & BLEU both \\
\midrule
\midrule
Cascaded        & 18.7      & 20.2      & 31.3     \\
\quad + punc    & 22.8      & 14.7      & 30.0     \\ \midrule
E2E YiTrans      & 18.0      & 19.1      & 29.8      \\
\quad + punc    & 21.8      & 13.7      & 28.2      \\
\bottomrule
\end{tabular}
}
\caption{\label{official_enja} Official results of our submitted \textbf{en-ja} ST systems on tst2022.}
\end{table}

\end{document}